\pdfoutput=1

\documentclass[11pt]{article}

\usepackage[]{EACL2023}

\usepackage{times}
\usepackage{latexsym}

\usepackage[T1]{fontenc}

\usepackage[utf8]{inputenc}

\usepackage{microtype}

\usepackage{graphicx}
\usepackage{booktabs}
\usepackage{caption}
\usepackage{multirow}
\usepackage{siunitx}
\usepackage{soul}
\usepackage{adjustbox}
\usepackage{tikz}
\newcommand*\circled[1]{\tikz[baseline=(char.base)]{
            \node[shape=circle,draw,inner sep=0.5pt] (char) {#1};}}

\title{Schema-Guided Semantic Accuracy: Faithfulness in Task-Oriented Dialogue Response Generation}

\author{Jinghong Chen \\
  Department of Engineering \\
  University of Cambridge \\
  United Kingdom \\
  \texttt{jc2124@cam.ac.uk} \\\And
  Weizhe Lin \\
  Department of Engineering \\
  University of Cambridge \\
  United Kingdom \\
  \texttt{wl356@cam.ac.uk} \\\And
  Bill Byrne \\
  Department of Engineering \\
  University of Cambridge \\
  United Kingdom \\
  \texttt{wjb31@cam.ac.uk} \\}

\begin{document}
\maketitle
\begin{abstract}

Ensuring that generated utterances are faithful to dialogue actions is crucial for Task-Oriented Dialogue Response Generation. Slot Error Rate (SER) only partially measures generation quality in that it solely assesses utterances generated from non-categorical slots whose values are expected to be reproduced exactly.   Utterances generated from categorical slots, which are more variable,  are not assessed by SER.  We propose Schema-Guided Semantic Accuracy (SGSAcc) to evaluate utterances generated from both categorical and non-categorical slots by recognizing textual entailment.   We show that SGSAcc can be applied to evaluate utterances generated from a wide range of dialogue actions in the Schema Guided Dialogue (SGD) dataset with good agreement with human judgment. We also identify a previously overlooked weakness in generating faithful utterances from categorical slots in unseen domains.  We show that prefix tuning applied to T5 generation can address this problem. We further build an ensemble of prefix-tuning and fine-tuning models that achieves the lowest SER reported and high SGSAcc on the SGD dataset. 

\end{abstract}

\section{Introduction}
\label{sec:Introduction}

Task-oriented dialogue response generation aims to generate accurate and fluent utterances from triplets of intent, slot, and values known as dialogue actions (See Fig.\ref{fig:SGSAcc procedure}). Ensuring that the generated utterances faithfully realize dialogue actions is crucial because misinformation can be costly in real-life applications. However, the lack of a complete, automatic faithfulness metric for task-oriented dialogue NLG has made assessing faithfulness difficult. Slot Error Rate (SER)~\cite{luong-etal-2015-effective},  the most widely used faithfulness metric currently, can only assess utterances generated from non-categorical slots 
whose values are expected to be reproduced exactly in generations by string matching, omitting utterances generated from categorical slots such as ``\texttt{(kids\_friendly, True)}''. However, dialogue actions with categorical slots are present in 28\% of the test instances in the Schema Guided Dialogue (SGD) dataset~\cite{DBLP:conf/aaai/RastogiZSGK20}, which is the largest dataset for multi-domain task-oriented dialogue system so far. Therefore, it is essential to include them for complete faithfulness evaluation. 
 
 To cover categorical slots in evaluation, we propose Schema-Guided Semantic Accuracy (SGSAcc) that examines semantic consistency rather than string overlap. We build upon Semantic Accuracy~\cite{DBLP:conf/inlg/DusekK20}, which evaluates faithfulness in table-to-text tasks by recognizing textual entailment (RTE). A natural language inference (NLI) model is used to check whether the premise (generated utterances) entails, contradicts, or is neutral to the hypothesis (dialogue actions).
Following their design, we first convert the dialogue actions into fluent sentences to serve as the hypothesis, so that the NLI model trained on free-running texts can perform RTE correctly without further fine-tuning. We name the converted sentences \emph{entailment reference} to emphasize their role in entailment checking.

We find that the original Semantic Accuracy cannot be directly applied on task-oriented response generation because it requires handcrafted templates for each dialogue action to produce entailment references. Although \citet{DBLP:conf/emnlp/KaleR20} have published templates for the SGD, they were designed for generation rather than evaluation. We find that Semantic Accuracy using these templates marks 25\% of the ground-truth utterances as unfaithful. To cover the 45 services in the SGD and the 225 service variations in the SGD-X without prohibitive labor, we propose a rule-based algorithm that constructs entailment references based on slot descriptions from service schema, which are provided in the SGD~\cite{DBLP:conf/emnlp/KaleR20} and other popular task-oriented dialogue datasets such as MultiWOZ 2.2~\cite{DBLP:journals/corr/abs-2007-12720}. We were able to design the rules efficiently within 20 working hours for the SGD (Appendix \ref{app:rule construction}). In addition, to help resolve co-references, which are prevalent in dialogues, we augment the premise (generation to be assessed) with previous dialogue turns and slot descriptions when needed. 
We verified that SGSAcc has good agreement with human judgments of faithfulness.

We applied SGSAcc to evaluate the best-performing published model in terms of SER on the SGD. We found a previously overlooked weakness in generating faithful utterances in domains not seen in training. To address this, we experimented with prefix-tuning (PT) \cite{DBLP:conf/acl/LiL20} which was reported to generalize better than fine-tuning (FT) on unseen data  \cite{DBLP:conf/acl/LiL20, DBLP:journals/corr/abs-2110-08329}. We found that PT significantly improved SGSAcc in unseen domains, whereas the FT model  achieved lower SER in comparison. Noting their complementary advantages, we used SGSAcc to implement a fidelity reranker similar to that from  \citet{DBLP:conf/coling/HarkousGS20} to select faithful generations from an ensemble of PT and FT models, which further improved SER and SGSAcc on the SGD.

Our contributions are summarised as follows:

(1) We propose SGSAcc, a faithfulness metric tailored to task-oriented dialogue systems that evaluate both categorical and non-categorical slots. 

(2)  We empirically show that prefix-tuning significantly improves faithfulness in unseen domains on the Schema Guided Dialogue dataset. 


(3) We build an ensemble of fine-tuning and prefix-tuning models using a SGSAcc-powered fidelity reranker, which significantly improves faithfulness in the NLG task of the SGD dataset.

\section{Schema-Guided Semantic Accuracy}
\label{sec:method:sg_sacc}

\begin{figure*}[h!]
    \centering
    \includegraphics[width=1.05\textwidth]{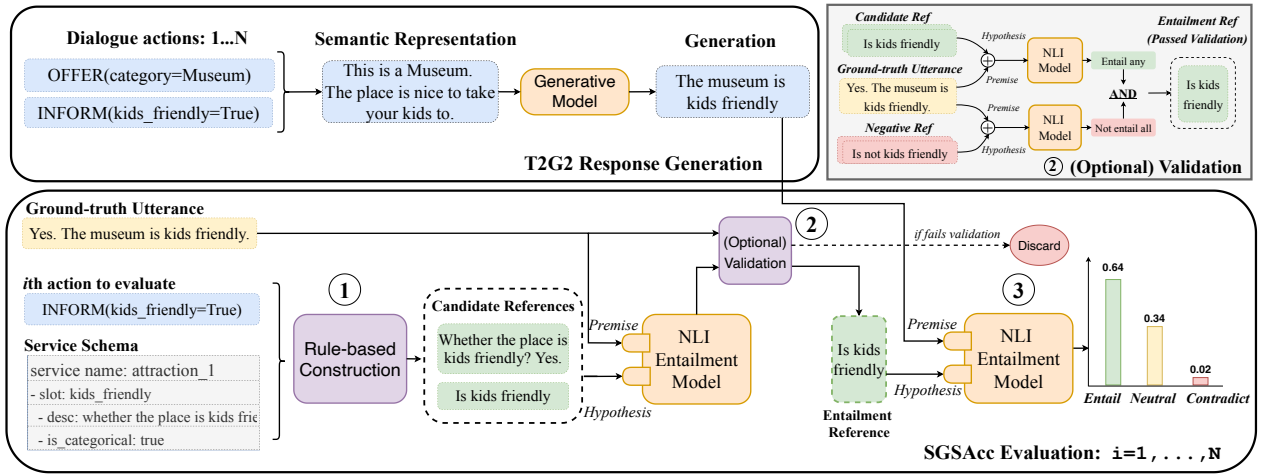}
    \caption{SGSAcc evaluation (Sec.~\ref{sec:method:sg_sacc}).
   (1) Construct multiple candidate references using predefined rules taking slot names and descriptions from schema; (2) The candidate with the highest entailment score is selected as the entailment reference. Optionally, the validation step checks that the ground-truth utterance entail at least one candidate reference and not entail any negative references constructed with substituted slot values. Otherwise, the instance is excluded from the evaluation; (3) RTE between the premise (the generated utterance, supplemented with previous dialogue turn and slot descriptions when needed) and the hypothesis (entailment reference). The same NLI model is used throughout SGSAcc evaluation. Here, the model outputs an entailment probability of 0.64, so the generation is considered faithful to the dialogue action.}
    \label{fig:SGSAcc procedure}
    \vspace{-0.4cm}
\end{figure*}

A generation is considered faithful if all dialogue actions are faithfully realized. For each dialogue action, faithfulness is evaluated through an NLI model, which checks whether the generated utterance entails the hypothesis text constructed from this dialogue action. The generation is considered faithful if ``entailment'' attains the highest probability amongst the three NLI output classes --- ``entailment'', ``neutral'', and ``contradiction'' (Fig.\ref{fig:SGSAcc procedure} \circled{3}). We use RoBERTa~\cite{DBLP:journals/corr/abs-1907-11692} without further fine-tuning as our NLI model (Appendix \ref{app:hyperpara}).

To obtain the entailment reference from the dialogue action, we first construct multiple candidate references using a set of rules designed for each type of dialogue action (Fig.\ref{fig:SGSAcc procedure} \circled{1}). For example, two candidate references are constructed from the dialogue action ``\texttt{INFORM(kids\_friendly=True)}'' by directly expanding the slot name into ``\textit{Is kids friendly.}'' and rephrasing the slot description from schema to an answered question ``\textit{Whether the place is kids friendly? Yes.}'' (See Appendix \ref{app:make hypos} for detailed rules). Then, using the ground-truth utterance as the premise, we select the candidate with the highest entailment score as the entailment reference. Optionally, we further check that the NLI model yields entailment for at least one of the candidate references and yields non-entailment for all negative references constructed from ``tampered'' dialogue actions with incorrect slot values (Fig.\ref{fig:SGSAcc procedure} \circled{2}). Instances that fail the check are excluded from evaluation. We call this optional checking \emph{validation}. Note that the validation step is agnostic to different generation models as it uses the ground-truth utterance as the premise, not the generated utterance. This ensures all models are evaluated equally on the same test set.
In our experiment on the SGD, only 3\% of the test instances fail the validation. We find that SGSAcc has better agreement with human on validated instances (see section \ref{sec:SGSAcc Evaluation}). We report SGSAcc with and without validation in our results (Table \ref{tab:SGD results}).

\begin{table}[h!]
\centering
\small
\begin{tabular}{ll}
\toprule
\textit{Dialogue action}    & inform (name = Queens)             \\
\textit{Entailment reference} & The name of the hair stylist is Queens.    \\ \midrule

\textit{ground-truth utterance} & What about Queens? \\
\textit{NLI} &  \multicolumn{1}{p{4.5cm}}{\textbf{neutral to} ``The name of the hair stylist is Queens.'' (neutral: 0.9) }
\\ \midrule
\textit{Previous turn}        & I want to book a hair cut. \\
\textit{Slot description}     & the name of the hair stylist            \\ 
\textit{Augmented premise} & \multicolumn{1}{p{4.5cm}}{I want to book a hair cut. the name of the hair stylist. What about Queens?} \\ 
\textit{NLI} & \multicolumn{1}{p{4.5cm}}{ \textbf{entails} ``The name of the hair stylist is Queens.'' (entailment: 0.7)}
\\
\bottomrule                     
\end{tabular}
\captionof{figure}{An example where the lack of dialogue context hinders faithfulness evaluation.}
\label{tab:coref example}
\vspace{-0.3cm}
\end{table}

\subsection{Supplying Dialogue Context to NLI}

We found that ground-truth utterances frequently refer to subjects that appeared in previous dialogue turns, which hinders proper entailment recognition. In the example in Figure \ref{tab:coref example}, without dialogue context, the NLI model outputs neutral with high confidence because it is unclear that ``Queens'' in the ground-truth utterance refers to a ``hair stylist''. To help capture the dialogue context, when the generated utterance alone fails to entail the entailment reference, we add the previous dialogue turn and the slot description to the premise and check for entailment again. This optional supplement is applied in the validation step and the final evaluation. 

\subsection{SGSAcc Evaluation}
\label{sec:SGSAcc Evaluation}

We conducted human evaluations with 3 crowd-sourced evaluators to verify that SGSAcc agrees with human judgment in faithfulness. We use the generations of the fine-tuned T5 model detailed in Sec.\ref{sec:generation models} for evaluation. With the validation step enabled, we randomly sample 200 utterance-dialogue action pairs which SGSAcc marks as faithful, and another 200 pairs which SGSAcc marks as unfaithful from the 9703 validated instances. Note that we use the ground-truth utterances rather than the generated utterances in the validation step, so the choice of generation model does not affect the set of validated instances. We asked the human evaluators to annotate whether the generated utterance faithfully verbalizes the dialogue action. The majority judgment was compared with the evaluation results of SGSAcc. Inter-annotator agreement is near-perfect, as indicated by a Fleiss' kappa of 0.93. 
SGSAcc with validation agrees with human judgments on 90.0\% of the examples, showing good agreement. We also 
randomly sample 25 examples from instances that fail the validation step and find that human agreement degrades to 76\%, suggesting that SGSAcc evaluation is more reliable on validated instances.

Compared with SER, SGSAcc provides significantly more complete evaluations by examining both non-categorical and categorical slots, increasing coverage from 66\% to 97\% (with validation) of the test set in the SGD. In addition, we also find that SGSAcc is consistent with SER in examining non-categorical slots, as SGSAcc yields the same result on 99.7\% of the test instances that can be evaluated by SER. 

Since SGSAcc uses schema information to construct candidate references, we also validated that SGSAcc is robust to different schema writing styles, as shown by the consistently high F1-score (>0.95) on distinguishing faithful and unfaithful utterances with the rephrased schema in the SGD-X extension~\cite{DBLP:journals/corr/abs-2110-06800} of SGD (See Table \ref{tab:SGSAcc robustness} in Appendix \ref{app:robustness writing styles}).

\begin{table*}[!h]
\centering
\small
\begin{tabular}{lrrrrrrrrrr} \toprule
\multicolumn{1}{c}{\multirow{2}{*}{Model}} & \multicolumn{1}{c}{\multirow{2}{*}{BLEU}} & \multicolumn{3}{c}{SER} & \multicolumn{3}{c}{SGSAcc(validated)} & \multicolumn{3}{c}{SGSAcc(all)} \\
\multicolumn{1}{c}{} & \multicolumn{1}{c}{} & \multicolumn{1}{c}{all} & seen & unseen & all & seen & unseen & all & seen & unseen \\ \midrule
CVAE~\cite{DBLP:conf/inlg/DuOPSNCVH20} & \textbf{43.0} & 24.0 & 9.00 & 27.0 & - & - & - & - & - & - \\
GPT2~\cite{DBLP:journals/corr/abs-2109-12211} & 20.5 & 0.90 & - & - & - & - & - &  - & -  & - \\
T2G2-T5~\cite{DBLP:journals/corr/abs-2109-12211} & 28.6 & 0.40 & 0.40 & 0.40 &  - & -  & - & - & - &-  \\   \midrule
FT-T5~(Ours) & 32.3 & 0.04 & 0.05 & 0.00 & 97.3 & 98.6 & 88.7 & 96.5 & 97.8 & 87.7 \\ 
PT-T5~(Ours) & 31.8 & 0.82 & 0.91 & 0.23 & 98.5 & 98.7 & 97.0 & 97.9 & 98.2 & 96.4 \\
PT+FT-T5~(Ours) & 32.1 & \textbf{0.02} & \textbf{0.02} & \textbf{0.00} & \textbf{98.9} & \textbf{99.5} & \textbf{97.7} & \textbf{98.5} & \textbf{99.1} & \textbf{97.1} \\ \bottomrule
\end{tabular}
\caption{Performance comparison between previous work on the SGD and ours. Best performance is in \textbf{bold}. SGSAcc scores with and without the validation step are close ($\pm 1\%$) and system rankings are preserved.}
\label{tab:SGD results}
\vspace{-0.5cm}
\end{table*}

\section{Faithful NLG with SGSAcc Ensembling}
\label{sec:generation models}

We now describe a Template-Guided (T2G2)~\cite{DBLP:conf/emnlp/KaleR20} NLG approach for the SGD dataset based on T5~\cite{DBLP:journals/jmlr/RaffelSRLNMZLL20}, Prefix-Tuning~\cite{DBLP:conf/acl/LiL20}, and SGSAcc ensembling, which shows that SGSAcc can also be used to improve the faithfulness of generation.

\noindent\textbf{Data Preprocessing.} We follow the T2G2 approach by \citet{DBLP:conf/emnlp/KaleR20} as shown in Fig.\ref{fig:SGSAcc procedure}. We first use predefined templates to turn each dialogue action into a natural sentence by substituting placeholders with the slot values. Then the templated sentences from each dialogue action are concatenated together to form the semantic representation (SR), which is fed into a generative model to generate utterances.

\noindent\textbf{Fine-Tuning T5 (FT-T5).}
 We fine-tuned a T5-small model \cite{DBLP:journals/jmlr/RaffelSRLNMZLL20} as our baseline generative model (details in Appendix \ref{app:hyperpara}).

\noindent\textbf{Prefix-Tuning T5 (PT-T5).} We also experimented with prefix-tuning, a technique that inserts trainable key-value pairs at each attention layer while fixing other parameters in the language model during training~\cite{DBLP:conf/acl/LiL20}. PT is reported to enhance generalization~\cite{DBLP:conf/acl/LiL20, DBLP:journals/corr/abs-2110-08329}. Details are given in Appendix \ref{app:hyperpara}.

\noindent\textbf{SGSAcc Ensemble (PT+FT-T5).} To further improve faithfulness, we adapt SGSAcc to implement a fidelity reranker~\cite{DBLP:conf/coling/HarkousGS20} that helps select faithful utterances from an ensemble of FT and PT models. We check whether the generated utterance entails any of the candidate references constructed from the dialogue action. If the NLI model outputs entailment for any of the candidates, the generation is considered faithful to the dialogue action. No labels are leaked as the NLI model has not seen the dataset and no ground-truth utterances are used to select entailment reference. For this SGSAcc ensemble, we first use beam search to decode the FT and PT models in parallel to obtain their respective best generations. Then, we use SGSAcc to assign fidelity scores to each generation, adding a score of 1 for each faithfully realized dialogue action and 0 otherwise. Finally, the generation that attains the highest fidelity score (or likelihood in case of a tie) is selected as the final generated utterance. The reranker targets for higher SGSAcc and we find that it also reduces SER.

\section{Experiments and Results}

We use the preprocessed Schema Guided Dialogue (SGD) dataset as part of the GEM benchmark~\cite{DBLP:conf/acl/SuDCJWLLZBS21} for NLG tasks. There are 164,982 turns for training and 10,000 turns for testing. There are 16 domains in total, 4 of which are only present in the test split. We use SER, SGSAcc, and BLEU~\cite{10.3115/1073083.1073135} as evaluation metrics.

We compare our system with previous work on the SGD dataset focusing on faithfulness in Table \ref{tab:SGD results}. In terms of SER, our FT model already surpasses the previous best-achieving model, T2G2-T5 from \citet{DBLP:conf/emnlp/KaleR20} (0.04\% v.s 0.40\%). Compared with its near-perfect SER, FT-T5 scores relatively low on SGSAcc, especially in domains not seen in training (89.1\%). This suggests that the model struggles to realize categorical slots in unseen domains. In comparison, PT-T5 yields significantly higher SGSAcc (89.1\% $\rightarrow$ 98.5\%) than FT-T5, faithfully realizing unseen categorical slots. Finally, the FT+PT ensemble with SGSAcc fidelity reranker combines the advantages of FT-T5 and PT-T5, achieving a substantial SGSAcc improvement in both seen and unseen domains (99.4\% and 98.8\%) while keeping SER close to zero (0.02\%). 

\section{Conclusion}

We present SGSAcc, a faithfulness metric for Task-Oriented Dialogue Response Generation. We applied SGSAcc to the SGD dataset and showed that it significantly increases coverage relative to SER and agrees well with human judgments. We also showed that prefix-tuning (PT) can improve faithfulness of generation in unseen domains compared with fine-tuning (FT). Finally, Our PT+FT ensemble with SGSAcc fidelity reranker establishes a new faithfulness benchmark on the SGD dataset.\footnote{Code wil be released at \url{https://github.com/EriChen0615/SchemaGuidedSemanticAccuracy}}

\newpage
\section{Limitation}
\label{app:limitation}

We note that since SGSAcc uses the slot descriptions in the schema to construct reference candidates, writing slot descriptions manually may be required to apply SGSAcc on datasets that do not come with schema. 
Although an increasing amount of research in Task-Oriented Dialogue Response Generation~\cite{DBLP:conf/inlg/DuOPSNCVH20, DBLP:conf/emnlp/KaleR20, DBLP:journals/corr/abs-2109-12211} have shown that introducing schema information can help improve NLG performance, only recent Task-Oriented Dialogue datasets such as the SGD and MultiWOZ 2.2 provide service schema. 

Although we have shown that SGSAcc is relatively robust to different schema writing styles (Appendix \ref{app:robustness writing styles}), certain phrasing of slot names and slot descriptions can affect NLI entailment recognition. For example, the use of double negatives can be difficult for the NLI model to recognize entailment as analyzed in Appendix \ref{app:SGSAcc bad cases}. To address this issue, dialogue system designers may consider using alias slot names and rephrasing descriptions in practice so that a 
fluent entailment reference can be readily constructed by pre-designed rules to facilitate entailment recognition. 

\section*{Ethics Statement}

We recognize that neural-based Task-Oriented Dialogue Response Generation can potentially convey misinformation that may result in loss or cause harm in real-life applications. Mitigating such risks is the primary goal of this research and our future work. However, extra care must be taken when neural-based response generation is deployed for mission-critical tasks such as ambulance service or crime reports to ensure that information is communicated clearly and accurately. We also note that using pretrained Natural Language Inference models without further fine-tuning can reduce the environmental impact compared with training on specific datasets for each task. The reduction is proportional to the size of the NLI model and the dataset, which can be quite significant in corporate-level applications.

\bibliography{anthology,custom}
\bibliographystyle{acl_natbib}

\appendix

\section{Robustness Against Schema Writing Styles}
\label{app:robustness writing styles}

\begin{table}[h!]
\center
\small
\begin{tabular}{llll}
\multicolumn{4}{c}{\textbf{SGSAcc robustness to writing style}} \\ \toprule
Dataset      & Precision & Recall  & F1-Score \\ \midrule
Original SGD & 95.8 & 96.8 & 0.963    \\
SGD-X v1     & 97.0   & 97.0   & 0.970     \\
SGD-X v2     & 96.3 & 95.8 & 0.960     \\
SGD-X v3     & 96.9 & 94.5 & 0.957    \\
SGD-X v4     & 96.9 & 94.7 & 0.958    \\
SGD-X v5     & 96.9 & 93.9 & 0.954   \\ \bottomrule
\end{tabular}
\caption{SGSAcc faithfulness classification results versus different schema writing styles in the SGD dataset and its SGD-X extension. Positive examples are pairs of ground-truth utterance and original dialogue action. Negative examples are constructed by substituting the slot values of dialogue actions.}
\label{tab:SGSAcc robustness}
\end{table}

Since SGSAcc uses the slot description in service schema to construct entailment reference, we check its robustness to different schema writing styles so that it can be used to evaluate a variety of services with heterogeneous interfaces. We use the SGD-X dataset~\cite{DBLP:journals/corr/abs-2110-06800}, which contains five versions of schema rephrased from the original SGD to test whether SGSAcc is sensitive to writing styles. For each version, we run the validation step and assess how well SGSAcc classify entailment reference and negative references with F1-score. Table \ref{tab:SGSAcc robustness} shows that SGSAcc can effectively recognize faithfulness for all versions of schema, demonstrating its robustness against different schema writing styles.

\section{Detailed Rules for Constructing Candidate References}
\label{app:make hypos}

The rules for constructing candidate references depend on the intent and the type of slot. Given a dialogue action $A$ with intent $d$, slot name $slot$, corresponding slot values $value1,...,valueN$, and the slot description $desc(slot)$ provided by the schema, we construct candidate references as follows:

\noindent For intents other than ``REQUEST'', ``GOODBYE'' and ``REQ\_MORE'', if $slot$ is \textbf{non-boolean}, the candidates are 

\begin{itemize}
    \item When there is only one slot value: ``\texttt{\{desc(slot)\} is \{value\}}'' and ``\texttt{\{slot\} is \{value\}}''
    \item When there are $N$ slot values: ``\texttt{\{desc(slot)\} are \{value1\}, ... and \{valueN\}}'' and  ``\texttt{\{slot\} are \{value1\}, ... and \{valueN\}}''
\end{itemize}

\noindent If $slot$ is \textbf{boolean}, the form of the candidates depend on the slot value.

\begin{itemize}
    \item If the value is \textbf{True}:  ``\texttt{\{desc(slot)\}? Yes.}'' and  ``\texttt{\{slot\}? Yes.}'' are used as candidates. In addition, if the slot name contains ``\texttt{has}'' or ``\texttt{have}'', then we append ``\texttt{Does \{slot\}}'' to the candidate list. If the slot name itself contains ``\texttt{is}'', then ``\texttt{\{slot\}}'' is added as a candidate. Otherwise, we prepend ``\texttt{has}'', ``\texttt{have}'', ``\texttt{is}'' to the slot name to form extra candidates. 
    \item If the value is \textbf{False}: ``\texttt{\{desc(slot)\}? No.}'' and ``\texttt{\{slot\}? No.}'' are used as candidates. We further construct other candidates similarly as the true case, except that we prepend the negative forms ``\texttt{has not}'', ``\texttt{have not}'' and ``\texttt{is not}'' to ``\texttt{\{slot\}}''. Note that when ``\texttt{is}'' is present in the slot name, we substitute it with  ``\texttt{is not}'' and add the substituted slot name as a candidate.
\end{itemize}

\noindent When intent is ``REQUEST'', ``GOODBYE'' or ``REQ\_MORE'', the candidates are
\begin{itemize}
    \item ``REQUEST'': ``\texttt{Request \{desc(slot)\}}'' and ``\texttt{Request \{slot\}}''.
    \item ``GOODBYE'':
    ``\texttt{Goodbye}: ``\texttt{Have a good day.}'', ``\texttt{Bye bye.}'' and \texttt{See you.}''.
    \item ``REQ\_MORE': ``\texttt{What else do you need?}'', ``\texttt{What else can I help you with?}'' and ``\texttt{Is there anything else?}''.
\end{itemize}

\section{Efficient rules construction and verification}
\label{app:rule construction}
We start with an initial guess of rules and then fine-tune them iteratively as follows. First, we analyze error patterns of examples that fail the validation step when the ground-truth utterance from the SGD is used as the candidate reference (see Sec.\ref{sec:method:sg_sacc}). Then, we revise the prompts based on the error pattern and verify the fix with the RoBERTa public endpoint\footnote{Available on https://huggingface.co/roberta-large-mnli} hosted on HuggingFace~\cite{DBLP:journals/corr/abs-1910-03771}. Finally, we perform the validation step again on the SGD and the SGD-X variants to obtain the F-measure of distinguishing faithful utterances from unfaithful ones. With the NLI model fixed, higher F-measures indicate more accurate entailment recognition due to improved prompts. In this way, we efficiently obtain the templates and rules with 3 iterations, and prompts are demonstrated to work well without the need to fix all corner cases. We believe that this makes SGSAcc a practical and useful metric which can be easily extended to other tasks and datasets.

\section{Qualitative analysis of instances that fail the validation step}
\label{app:SGSAcc bad cases}

We provide two instances that fail the validation step in Table \ref{tab:sg-sacc bad cases} for qualitative analysis.

\begin{table}[h!]
\small
\begin{tabular}{lp{1.5cm}p{2cm}p{2.5cm}}
\toprule
 & Dialog Act                           & ground-truth utterance   (Premise)                                                                                                   & Candidate References (Hypotheses)                                                                                                                                                                          \\ \midrule
1      & inform (is\_nonstop = False)         & I found an American Airlines flight taking   off at 4:10 pm with a layover and returning at 1:40 pm. The price is \$449 & Whether   the flight is a direct one? No. ||  is   nonstop? No.  || is not nonstop ||   Flights is not nonstop                                                                                  \\  \rule{0pt}{0.1ex} \\
2      & inform (additional\_luggage = False) & You would like 2 tickets from Anaheim to Las Vegas on a bus departing on March   5th at 6:40 am?                      & Whether to carry excess baggage in the bus? No. || additional luggage? No. || Has no additional luggage || Is not additional luggage || Does not additional luggage                      \\ 
\bottomrule
\end{tabular}
\caption{Two instances that fail the SGSAcc validation step as the NLI model does not recognise entailment between the ground-truth utterance and any candidate reference. The candidate references constructed by different rules are separated by ``||''.}
\label{tab:sg-sacc bad cases}
\end{table}

Multiple rules are at work to construct multiple candidate references, but none of them yields entailment when the ground-truth utterance is used as the premise. The two examples correspond to two types of errors. (1) \emph{Complex rephrasing}: In the first example, indirect flight is rephrased as ``\emph{flight with a layover}'', and the model fails to recognize it as semantically equivalent to the double negative ``\emph{not non-stop}''. Although humans can tell that the reference indeed entails the hypotheses, the model-based entailment model might have failed to overcome the complexity of rephrasing and double negatives. (2) \emph{Implicit/Noisy ground-truth utterance}: The ground-truth utterance itself could be problematic or implicit about a certain slot realization. In the second example, the ground-truth utterance does not mention whether additional luggage is needed. Therefore, it fails to entail all candidate references, which explicitly include ``\texttt{additional baggage}'' as the subject. This type of errors  should be considered as noise in the dataset. 

\section{Model Configuration and Training Parameters}
\label{app:hyperpara}

For the NLI model, we use the ``\texttt{roberta-large-mnli}'' checkpoint from HuggingFace Transformers library~\cite{DBLP:journals/corr/abs-1910-03771} without further fine-tuning. 

Both the FT model and the PT model use the ``\texttt{t5-small}'' checkpoint from the Transformers library~\cite{DBLP:journals/corr/abs-1910-03771}. Both are trained with AdamW optimizer using a fixed learning rate of 1e-4, batch size of 8 for 3 epochs. We found that using smaller batch size and stopping at 3 epochs improves the model's performance, especially in unseen domains. Each epoch takes around 26 minutes and 20 minutes to train on a A100 GPU for the FT model and the PT model, respectively. 

For the PT model, the prefix length is set to 10 (the number of key-value pairs inserted to each attention layer). We use three separate MLPs to generate prefixes for encoder self-attention, decoder masked attention and decoder cross-attention, respectively similar to CONTROL-PREFIXES~\cite{DBLP:journals/corr/abs-2110-08329}. Each MLP is a two-layer fully connected network with the two hidden dimensions being 512 and 384. In this setting, there are 7.7M trainable parameters out of the total 68.2M parameters.




\end{document}